%% file: sections/main.tex
\documentclass[dvipsnames]{article} %
\PassOptionsToPackage{hyperfootnotes=false, pageanchor=false}{hyperref}

\usepackage{colm2024_conference}

\usepackage{booktabs}
\usepackage{enumitem}
\usepackage{wrapfig}
\usepackage{algorithmic}
\usepackage{graphicx}
\usepackage{subcaption}
\usepackage[misc]{ifsym}
\usepackage{algorithm}

\usepackage{microtype}
\usepackage{amsmath}
\usepackage{colortbl}
\usepackage[utf8]{inputenc}
\usepackage[T1]{fontenc}
\definecolor{lightgray}{rgb}{0.9,0.9,0.9}
\usepackage{caption}
\usepackage{subcaption}
\usepackage{setspace}
\usepackage{url}
\usepackage{multirow}
\usepackage{tabularx}
\usepackage{blindtext}
\usepackage{pgfplots}
\pgfplotsset{compat=1.18} 
\usepackage{tikz}
\usetikzlibrary{er,positioning,bayesnet}
\usepackage{makecell}
\usepackage{siunitx}
\usepackage{nicefrac}
\usepackage{listings}
\usepackage[raster,skins, most]{tcolorbox} %
\usepackage{xltabular}
\usepackage{adjustbox}
\usepackage{xurl}
\usepackage{rotating}
\usepackage[normalem]{ulem}
\usepackage{xcolor}

\usepackage{fontawesome}

\useunder{\uline}{\ul}{}

\usepackage{cleveref}

\input{math_commands.tex}

\hypersetup{
    breaklinks,
    citecolor=uclablue_old,
    colorlinks=true,
    hyperfootnotes=false, 
    pageanchor=false,     
}

\newcommand*\justify{%
  \fontdimen2\font=0.4em
  \fontdimen3\font=0.2em
  \fontdimen4\font=0.1em
  \fontdimen7\font=0.1em
  \hyphenchar\font=`\-
}

\renewcommand{\texttt}[1]{%
  \begingroup
  \ttfamily
  \begingroup\lccode`~=`/\lowercase{\endgroup\def~}{/\discretionary{}{}{}}%
  \begingroup\lccode`~=`[\lowercase{\endgroup\def~}{[\discretionary{}{}{}}%
  \begingroup\lccode`~=`.\lowercase{\endgroup\def~}{.\discretionary{}{}{}}%
  \catcode`/=\active\catcode`[=\active\catcode`.=\active
  \justify\scantokens{#1\noexpand}%
  \endgroup
}

\usepackage{makecell}
\usetikzlibrary{tikzmark}
\makeatletter
\newcommand*\myfontsize{%
  \@setfontsize\myfontsize{7}{8}%
}
\makeatother

\definecolor{uclablue}{RGB}{159, 195, 224}

\definecolor{uclagold}{RGB}{255, 240, 180}

\definecolor{aliceblue}{RGB}{255, 238, 241}

\definecolor{cadmiumgreen}{rgb}{0.0, 0.42, 0.24}

\definecolor{myred}{rgb}{0.7, 0.3, 0.0}
\definecolor{myblue}{rgb}{0.2, 0.3, 0.6}
\definecolor{babygreen}{rgb}{0.85, 0.97, 0.85}

\definecolor{purple1}{RGB}{126, 107, 196}
\definecolor{purple2}{RGB}{199, 158, 207}
\definecolor{purple3}{RGB}{214, 200, 255}
\definecolor{purple4}{RGB}{254, 240, 255}

\definecolor{deepblue}{RGB}{48, 58, 82}

\newcommand{\symboletongyi}{\raisebox{0pt}{~\includegraphics[scale=0.012]{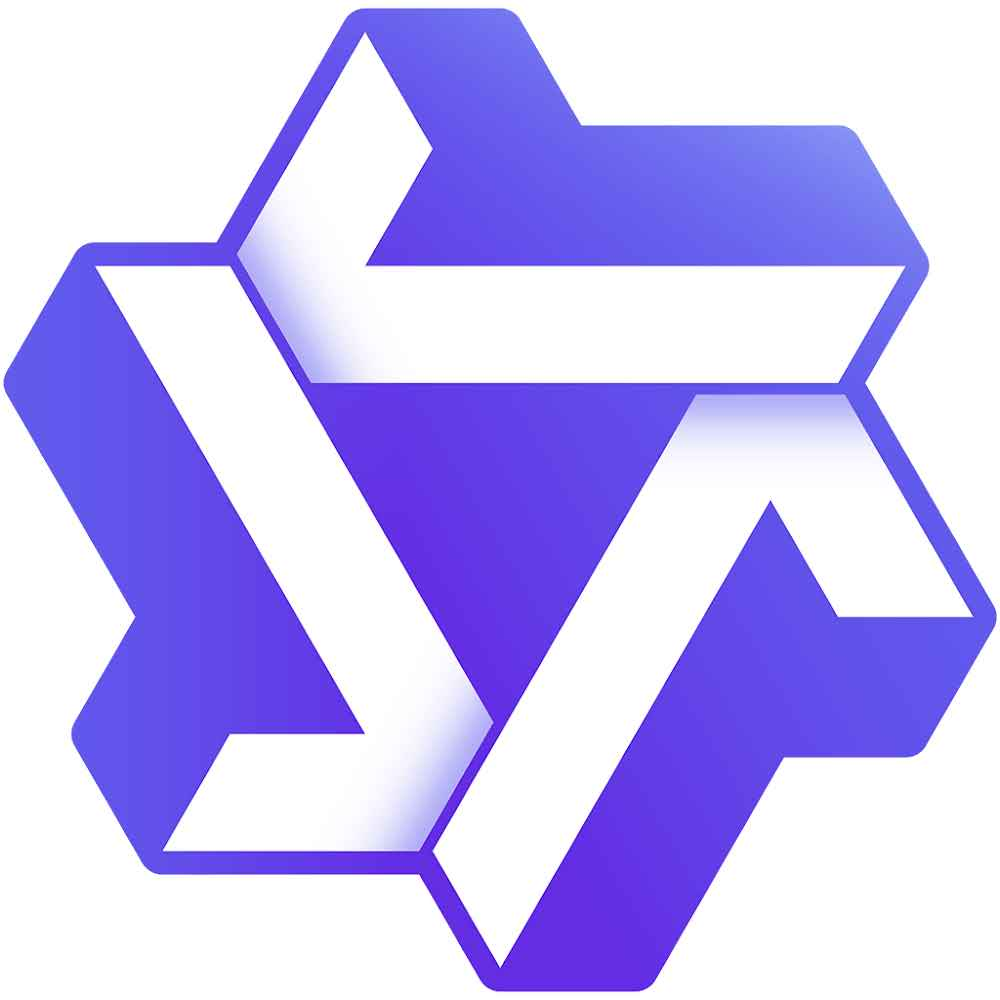}}~}

\definecolor{deepPurple}{HTML}{330066}

\definecolor{uclablue_old}{rgb}{0.15, 0.45, 0.68}

\newtcolorbox{mybox}[2][]
  {colback = black!5!white, colframe = black!75!black, fonttitle = \bfseries,
    colbacktitle = black!100!black, enhanced, before upper={\fontsize{8}{11}\obeyspaces\obeylines\selectfont}, fontupper=\selectfont,
    attach boxed title to top left={yshift=-2.2mm,xshift=4mm},
    title=#2,#1}

\title{ESPO: Early-Stopping Proximal Policy Optimization}

\author{%
Zihang Li$^{1,2}$, Rui Zhou$^{2}$, Yingcheng Shi$^{1}$$^{\dagger}$, Wenhan Yu$^{2}$, Zhewen Tan$^{2}$, Zixiang Liu$^{2}$, Zeming Li$^{2}$, Binhua Li$^{1}$, Yongbin Li$^{1}$, Tong Yang$^{2}$$^{\dagger}$, Jieping Ye$^{1}$%
  \medskip\\
  {\fontsize{10pt}{11pt}\selectfont
$^{1}$Tongyi Lab\symboletongyi, Alibaba Group  \quad $^{2}$Peking University}}

\usepackage{pifont}

\makeatletter
\def\@maketitle{\vbox{\hsize\textwidth
  {\centering {\Large\bf \@title\par}}
  \vskip 0.2cm
  {\centering \@author \par}%
  \vskip 0.3in minus 0.1in
}%
\thispagestyle{firstpage}}
\makeatother
\begin{document}

\maketitle
\makeatletter
\begingroup
  \let\@currentHref\@empty
  \renewcommand\@makefnmark{}%
  \footnotetext[1]{$^\dagger$Corresponding author.}
\endgroup
\makeatother

\begin{abstract}
\input{sections/0_abstract}

\end{abstract}

\input{sections/1_introduction}
\input{sections/2_related_work}

\input{sections/3_preliminaries}
\input{sections/4_method}

\input{sections/5_experiments}
\input{sections/6_ablation}

\input{sections/7_conclusion}

\bibliographystyle{plainnat}
\bibliography{references}

\newpage
\appendix
\input{sections/A_appendix}


\end{document}

%% file: math_commands.tex
\usepackage{amsthm}
\usepackage{mathtools}

\newcommand{\pitheta}{\pi_\theta}          
\newcommand{\Vphi}{V_\phi}                 

\newcommand{\st}{s_t}                      
\newcommand{\Tmax}{T_{\max}}               

\newcommand{\rfail}{r_{\mathrm{fail}}}     

\newcommand{\alphasm}{\alpha}              
\newcommand{\betath}{\beta}               

\newcommand{\clip}{\mathrm{clip}}



%% file: sections/0_abstract.tex
When a large language model under reinforcement learning commits a wrong reasoning
step early in a trajectory, standard algorithms force it to keep generating until
the maximum horizon, spending compute on tokens that never receive positive reward
and polluting advantage estimates with post-failure noise.
We propose \textbf{ESPO} (\textbf{E}arly-\textbf{S}topping Proximal \textbf{P}olicy
\textbf{O}ptimization), which detects trajectory failure on-the-fly and terminates
rollouts early.
At each generation step, ESPO computes a surrogate regret
using only the logits already computed during sampling,
and terminates when the smoothed cumulative regret significantly exceeds its estimated values.
Truncated trajectories are treated as absorbing failure states with a terminal
reward, concentrating negative temporal-difference (TD) errors near the detected failure
step without any additional reward model or human annotation.
On DeepSeek-R1-Distill-Qwen-7B trained for mathematical reasoning, ESPO surpasses PPO on
AIME~2024 (46.28\% vs.\ 45.25\%), AMC~2023 (85.83\% vs.\ 82.94\%),
and MATH-500 (87.42\% vs.\ 85.43\%),
while saving more than 20\% rollout tokens cumulatively.

%% file: sections/1_introduction.tex
\section{Introduction}
\label{sec:intro}

Reinforcement learning (RL) for large language models (LLMs) has emerged as a dominant paradigm for improving reasoning ability, particularly in multi-step problem solving tasks \citep{ouyang2022training}. Process-supervised approaches further demonstrate that RL can substantially enhance chain-of-thought reasoning performance \citep{lightman2023let}. And alignment-oriented RLHF pipelines have become central to modern LLM post-training \citep{casper2023open}. In these settings, models must generate trajectories of hundreds to thousands of tokens, each conditioned on the entire preceding context. This long-horizon generation process introduces significant challenges in credit assignment and training efficiency \citep{shao2024deepseekmath}. When the model takes an inappropriate reasoning step at $t^*$---for example, misidentifying a mathematical operation, straying from the topic in a processing paper, or branching down an incorrect proof path---the subsequent trajectory cannot recover. The eventual reward is zero or negative, yet standard policy gradient algorithms force the policy to continue generating until the end or the fixed rollout cap $T_{\max}$. These post-failure tokens receive no positive reward but are included in the advantage estimates alongside tokens from successful trajectories, introducing noisy gradient directions that misdirect the learning process away from the actual failure mode. This leads to wasted computation on sequences that will never improve the policy. We call this the \emph{rollout continuation problem}.

Existing works address adjacent problems but leave rollout continuation waste largely unsolved. Process reward models \citep{lightman2023let} provide step-level feedback but require extensive human annotation of intermediate reasoning steps. Methods such as GRPO \citep{shao2024deepseekmath} and DAPO \citep{yu2025dapo} improve credit assignment through group-normalized advantages and clipped importance weights. However, these methods still exhaust the full horizon for every trajectory. Learned termination approaches, such as Option-Critic \citep{bacon2017option}, introduce a dedicated termination module that must be trained alongside the main policy, adding model complexity and a separate optimization objective. None of these methods detects failure on-the-fly using only the signals already produced by the actor and critic during a standard PPO \citep{schulman2017proximal} step.

We introduce \textbf{ESPO} (\textbf{E}arly-\textbf{S}topping Proximal \textbf{P}olicy \textbf{O}ptimization), a lightweight rollout termination mechanism that reuses the policy's existing logit vector and the critic's value estimate with negligible additional computation. The core insight is that a policy in a high-regret, low-value state is less likely to recover: the gap between the action the policy \emph{would have taken} greedily and the action it \emph{did take} diverges sharply at failure points, while the critic simultaneously assigns low remaining value. ESPO formalizes this intuition through the following components:
\begin{enumerate}
    \item \textbf{Per-step surrogate regret.} 
    A regret signal computed from the logit gap at each timestep, capturing deviation from the greedy action.
    
    \item \textbf{EMA normalization.} 
    An exponential moving average (EMA) scheme keeps the regret signal scale-comparable to the critic's value estimate throughout training.
    
    \item \textbf{Value-gated stopping criterion.} 
    A termination rule stops the rollout when the normalized cumulative regret significantly exceeds the estimated value.
    
    \item \textbf{Terminal failure penalty.} 
    Truncation is treated as an absorbing failure state, producing a concentrated negative TD-error at the stopping step without introducing the non-stationary bias associated with per-step reward shaping.
\end{enumerate}

Experiments on 1.5B and 7B scales show that ESPO consistently outperforms both PPO and DAPO across almost all benchmarks while consuming much fewer rollout tokens. For example, at the 1.5B scale, ESPO achieves an average accuracy of 59.09\% across all three benchmarks, outperforming both PPO (57.03\%) and DAPO (58.29\%), while consuming 927.96M cumulative tokens---significantly less than DAPO (1223.96M, $-$24\%) and PPO (1069.66M, $-$13\%). The ablation against random truncation (Variant F in \Cref{tab:ablation}), which matches ESPO's stopping rate but ignores policy confidence and value estimates, scores only 42.4\% on AIME 2024 despite a similar average rollout length---confirming that the improvement stems from \emph{where} trajectories are truncated, rather than the reduced token budget alone.

\begin{figure*}[t]
    \centering
    \begin{minipage}[t]{0.49\textwidth}
        \centering
        \includegraphics[width=\linewidth]{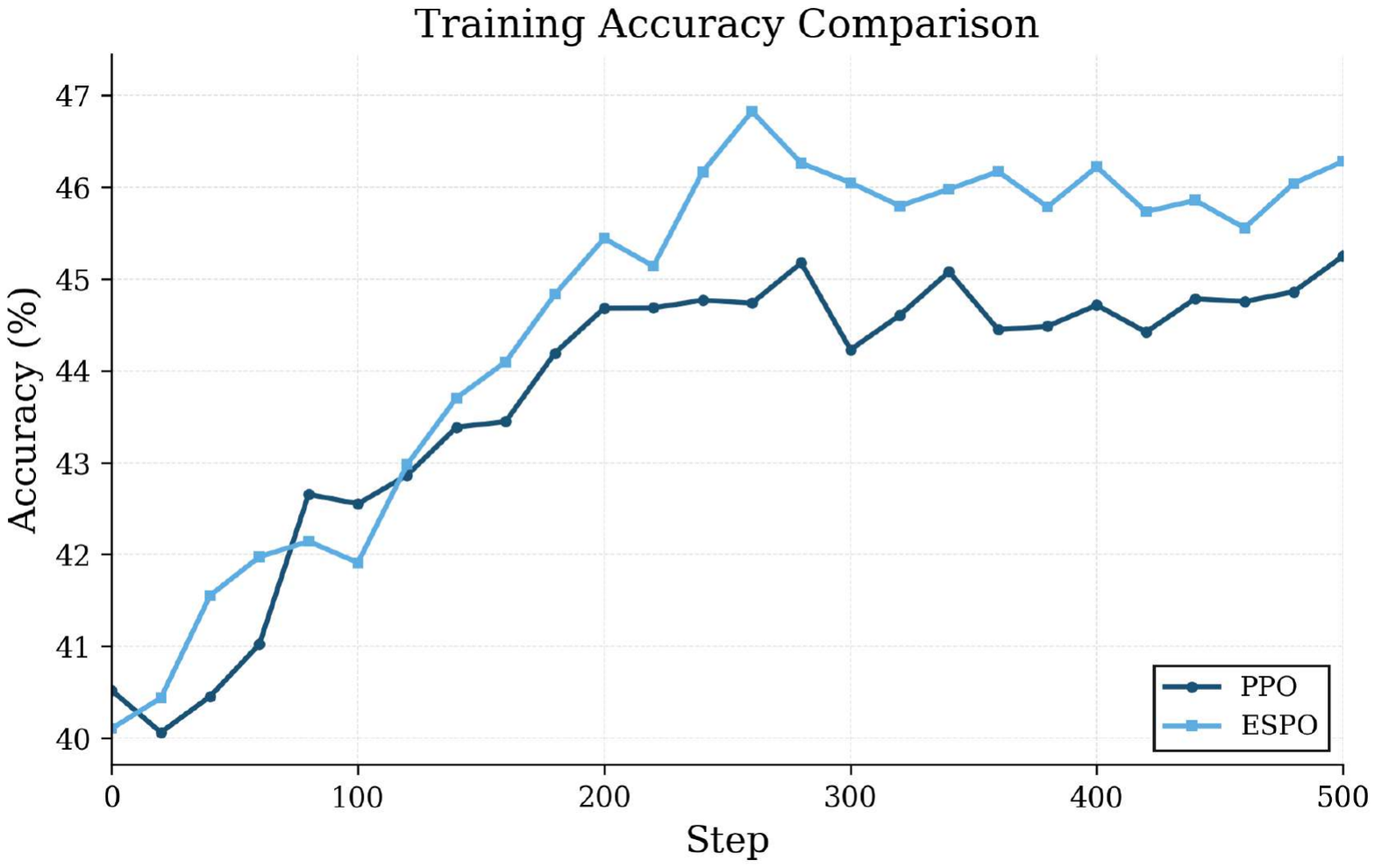}
        \subcaption{Accuracy}
    \end{minipage}
    \hfill
    \begin{minipage}[t]{0.49\textwidth}
        \centering
        \includegraphics[width=\linewidth]{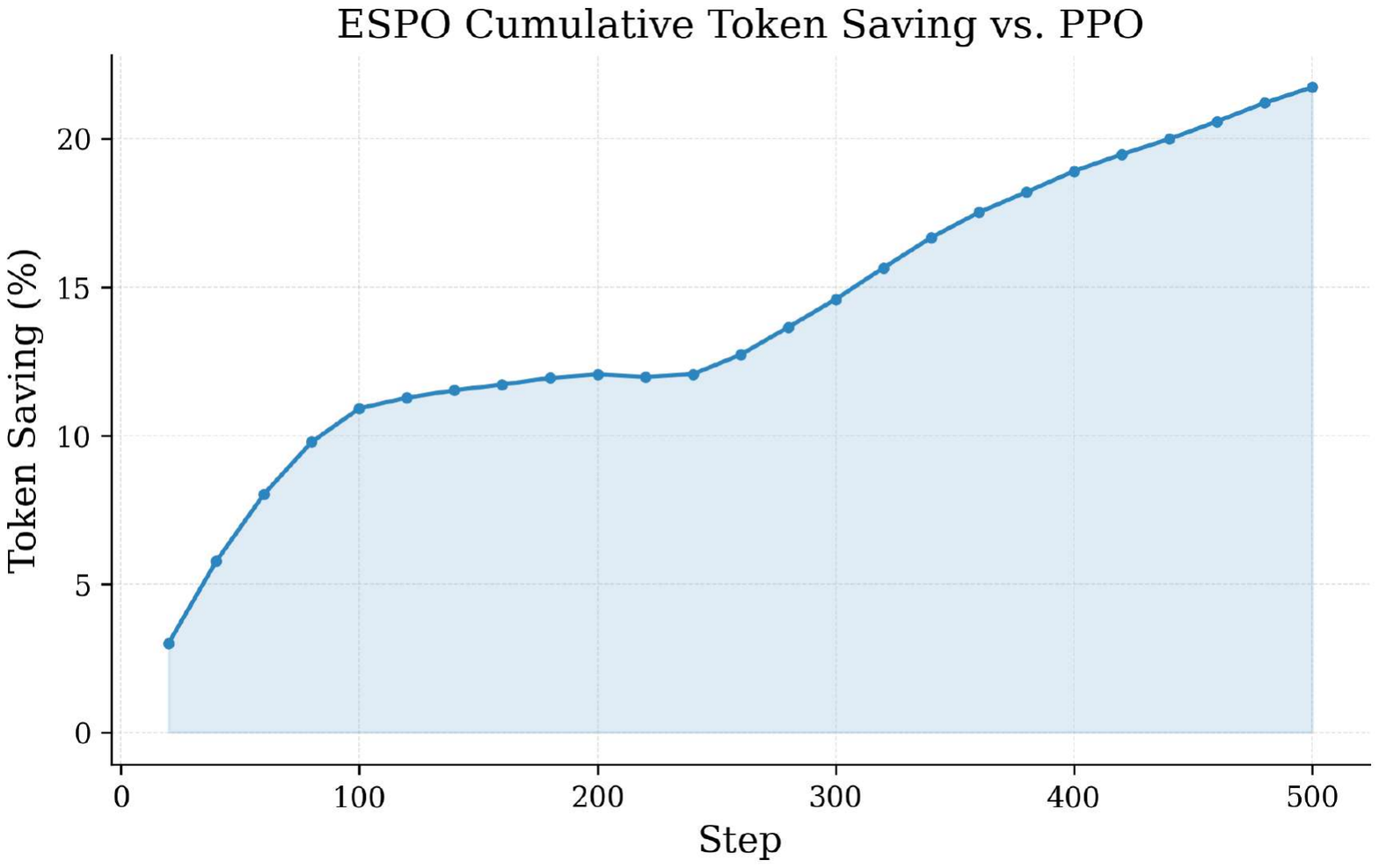}
        \subcaption{Token Saving}
    \end{minipage}
    \caption{%
        \textbf{ESPO surpasses PPO on AIME~2024 at lower token cost (DeepSeek-R1-Distill-Qwen-7B).}
        \textbf{Left:} AIME~2024 avg@32 (\%) vs.\ gradient update steps.
        ESPO surpasses PPO earlier and maintains the lead through training.
        \textbf{Right:} It records ESPO's cumulative rollout token saving during training compared with PPO.
        (tokens per step differ because ESPO truncates failing trajectories).
        All methods use identical prompts, reward functions, and evaluation protocol.
    }
    \label{fig:hero}
\end{figure*}




%% file: sections/2_related_work.tex
\section{Related Work}
\label{sec:related}

\paragraph{PPO-based LLM alignment and reasoning.}
Proximal Policy Optimization (PPO) \citep{schulman2017proximal} is the
standard backbone for RLHF in LLM alignment
\citep{ouyang2022training, bai2022constitutional, touvron2023llama},
due to its stable clipped objective and implicit trust-region control.
Recent work improves credit assignment and stability for long-horizon reasoning.
GRPO \citep{shao2024deepseekmath} reduces variance via group-wise advantage normalization,
removing the need for a learned baseline.
DAPO \citep{yu2025dapo} introduces dynamic sampling and advantage clipping.
GSPO \citep{zhang2025gspo} further stabilizes training through gradient-level
regularization and structured policy updates.
Despite these refinements, all PPO-style methods share a key inefficiency:
they roll out trajectories to a fixed horizon $T_{\max}$ even after
irrecoverable errors.
ESPO is orthogonal to these methods—its stopping criterion can be layered on top of any PPO-style advantage estimator.

\paragraph{Process reward models and step-level credit assignment.}
Sparse outcome rewards are a major bottleneck in multi-step reasoning.
Process reward models (PRMs) \citep{lightman2023lets}
provide dense step-level supervision but require costly human annotation.
Outcome reward models (ORMs) \citep{cobbe2021training,uesato2022solving} scale better but still
depend on full trajectories.
Preference-based methods such as DPO \citep{rafailov2023direct}
avoid online RL but forgo on-policy exploration and do not extend naturally
to interactive agentic settings.
In contrast, ESPO requires no step-level annotation and no separate
reward model: it derives its failure signal directly from the actor's
logit vector and the critic's value estimate.

\paragraph{Learned termination and the options framework.}
Termination has been studied in RL via the
Option-Critic framework \citep{bacon2017option}, which learns
both the option policy and a dedicated termination function in a joint end-to-end framework.
Pardo et al.~\citep{pardo2018time} show that mismatched time limits introduce
bias in value estimation, motivating a distinction between truncation and
natural terminal states.
ESPO adopts this perspective: it maps forced truncations to absorbing failure
states rather than mid-episode transitions.
Unlike prior work, ESPO requires no learned termination module—it reuses
signals already computed in a standard PPO forward pass.

\paragraph{Inference-time early stopping for reasoning models.}
A parallel line of work targets the overthinking problem at inference time
rather than during training. ESTAR~\citep{wang2026estar} introduces a token-aware
early-stopping criterion that monitors the model's evolving answer to terminate
chain-of-thought generation once a stable prediction emerges. TERMINATOR~\citep{nagle2026terminator}
instead learns optimal exit points by training a predictor on the first-arrival
positions of the final answer, achieving 14--55\% length reduction on reasoning
benchmarks. Both approaches operate purely at inference and assume a fixed,
already-trained policy. ESPO is complementary: it truncates trajectories
\emph{during RL training} to remove post-failure noise from the policy gradient,
yielding a policy whose own rollouts are shorter and more accurate at inference time
without requiring any auxiliary inference-time controller.

\paragraph{Efficiency-oriented RL for reasoning.}
Several recent methods address compute inefficiency directly within the RL
post-training stage. DRPO~\citep{li2026drpo} identifies that GRPO's group-relative
advantage can assign negative advantages to correct-but-long rollouts when length
penalties are introduced, and decouples the length-based learning signal of correct
and incorrect rollouts to mitigate this misclassification. Latent-GRPO~\citep{deng2026latentgrpo}
takes a more structural approach, performing policy optimization in a continuous
latent reasoning space rather than over explicit token chains, thereby compressing
reasoning into shorter trajectories. ESPO differs from both in mechanism and scope:
rather than reshaping the reward (DRPO) or moving reasoning into latent space
(Latent-GRPO), ESPO leaves the reward and the token-level action space unchanged,
and instead intervenes at the rollout-collection level by detecting on-the-fly when
a trajectory has entered an irrecoverable failure mode. This makes ESPO orthogonal
and composable.

%% file: sections/3_preliminaries.tex
\section{Preliminaries}
\label{sec:prelim}

\paragraph{Token-level RL for LLM generation.}
We formulate autoregressive decoding as a finite-horizon Markov decision process.
Given an input prompt $x \sim \mathcal{D}$, the state at step $t$ is
$s_t = (x, y_{<t})$, namely the prompt together with the previously generated tokens.
The action $a_t \in \mathcal{V}$ is the next token selected from the vocabulary,
and the transition is deterministic:
$s_{t+1} = (x, y_{<t}, a_t)$.
An episode terminates when the model emits an end-of-sequence token or reaches
the horizon $T_{\max}$.
Following standard outcome-supervised RL for reasoning tasks, we consider sparse rewards,
\begin{equation}
    r_t = 0 \quad (t < T), \qquad r_T = R(x, y_{1:T}),
\end{equation}
where $R(x, y_{1:T})$ denotes the final task reward, e.g., a binary correctness score.

\paragraph{Actor-critic PPO.}
Let $\pi_\theta(a_t \mid s_t)$ denote the policy and $V_\phi(s_t)$ the value of critic.
PPO updates $\pi_\theta$ by maximizing the clipped surrogate objective
\begin{equation}
\mathcal{L}_{\mathrm{PPO}}(\theta)
=
\mathbb{E}_{t}\left[
\min\left(
\rho_t \hat{A}_t,
\clip(\rho_t, 1-\epsilon_{\mathrm{ppo}}, 1+\epsilon_{\mathrm{ppo}})\hat{A}_t\right)
\right],
\label{eq:ppo}
\end{equation}
where
\begin{equation}
    \rho_t
    =
    \frac{\pi_\theta(a_t \mid s_t)}{\pi_{\mathrm{old}}(a_t \mid s_t)}
\end{equation}
is the importance ratio.
Advantages are estimated with generalized advantage estimation (GAE):
\begin{equation}
    \delta_t = r_t + \gamma V_\phi(s_{t+1}) - V_\phi(s_t),
\end{equation}
\begin{equation}
    \hat{A}_t = \sum_{l=0}^{T-t-1} (\gamma \lambda)^l \delta_{t+l}.
\label{eq:gae}
\end{equation}
This actor-critic formulation is important for our method because the critic
provides a state-dependent estimate of remaining return, while GAE propagates
the effect of early termination backward to preceding tokens.

%% file: sections/4_method.tex
\section{ESPO: Early-Stopping Proximal Policy Optimization}
\label{sec:method}

\paragraph{Overview.}
ESPO modifies \emph{rollout collection} rather than the PPO objective itself.
At each decoding step, it computes a cheap token-level deviation signal from the
policy logits, smooths this signal over time, and compares it against a
value-dependent threshold derived from the critic.
Once the threshold is exceeded, the trajectory is terminated and mapped to an
absorbing failure transition with a terminal penalty.
PPO and GAE are then applied to the truncated trajectory in the usual way.
In this sense, ESPO can be viewed as PPO on an augmented episodic MDP whose
termination rule is induced online during generation.

\subsection{Stepwise deviation signal}
\label{sec:stepwise_signal}

At state $s_t$, let $a_t \sim \pi_\theta(\cdot \mid s_t)$ be the sampled token.
We define the stepwise deviation signal (also called the regret value) as
\begin{equation}
    g_t
    =
    \max_{a \in \mathcal{V}} \log \pi_\theta(a \mid s_t)
    -
    \log \pi_\theta(a_t \mid s_t).
\label{eq:step_gap}
\end{equation}
By construction, $g_t \ge 0$.
The quantity is small when the sampled token is close to the policy mode,
and increases when sampling deviates from the policy's most preferred action.
Importantly, $g_t$ is obtained directly from the logits already computed for decoding
and therefore introduces negligible additional computation cost.

\subsection{Normalized cumulative stopping statistic}
\label{sec:normalized_stat}

Because the scale of $g_t$ changes during training, we first normalize it with
running batch statistics.
Let $\mu_g$ and $\sigma_g^2$ denote an exponential moving average of the batch mean
and variance of $g_t$:
\begin{align}
    \mu_g &\leftarrow \alpha_{\mathrm{ema}} \mu_g + (1-\alpha_{\mathrm{ema}})\,\overline{g}_{\mathcal{B}},
    \\
    \sigma_g^2 &\leftarrow \alpha_{\mathrm{ema}} \sigma_g^2 + (1-\alpha_{\mathrm{ema}})\,\mathrm{Var}(g_{\mathcal{B}}),
\end{align}
where $\mathcal{B}$ denotes the current rollout batch.
Crucially, to ensure causal correctness and prevent future information within the current rollout from leaking into the termination decision, $\mu_g$ and $\sigma_g^2$ are updated only at the boundary of each training batch.
Using these strictly frozen batch-level statistics during generation, we define the normalized signal
\begin{equation}
    \tilde{g}_t
    =
    \operatorname{clip}\left(
    \frac{g_t - \mu_g}{\sqrt{\sigma_g^2 + \delta}},
    -c, c
    \right),
\label{eq:normalized_gap}
\end{equation}
where $\delta > 0$ is a numerical stabilizer and $c$ is a clipping bound.
We then accumulate the normalized signal within each trajectory using
\begin{equation}
    z_t = \alpha_{\mathrm{s}} z_{t-1} + (1-\alpha_{\mathrm{s}})\tilde{g}_t,
    \qquad z_0 = 0.
\label{eq:smoothed_score}
\end{equation}
The scalar $z_t$ (also called the cumulative regret value) serves as the stopping statistic used by ESPO.

\subsection{Value-gated early termination}
\label{sec:value_gated_stop}

ESPO triggers early termination at the current step $t$ if the cumulative deviation satisfies
\begin{equation}
    z_t > \beta \cdot \max\bigl(V_\phi(s_t),\;\varepsilon\bigr).
\label{eq:stop_rule}
\end{equation}
$\varepsilon$ acts as the threshold. The gating has a simple interpretation: states with high predicted future return
are granted larger tolerance, whereas states with low predicted return are terminated
after a smaller amount of accumulated deviation. This also to some extent prevent the correct but non-mode token (with high regret) from being wrongly terminated.

To maintain a stable stopping frequency throughout training, the threshold multiplier
can be adjusted by a proportional controller,
\begin{equation}
    \beta
    \leftarrow
    \operatorname{clip}\!\left(
    \beta + \eta_\beta (\hat{\rho}_{\mathrm{stop}} - \tau),
    \beta_{\min}, \beta_{\max}
    \right),
\label{eq:beta_update}
\end{equation}
where $\hat{\rho}_{\mathrm{stop}}$ is the empirical stop rate and $\tau$ is the
target rate.
In practice, evaluating Equation~\eqref{eq:stop_rule} immediately would induce spurious early terminations due to uncalibrated value baselines from the randomly initialized critic. Therefore, we disable the stopping rule during an adaptive critic warmup period, which dynamically concludes once the critic exhibits stable learning dynamics (e.g., when validation loss improvement drops below a specified tolerance continuously). More details about warmup are stated in Appendix~\ref{app:warmup}. Following this burn-in, $\beta$ is linearly annealed from a conservative upper bound down to its target regime, safely transitioning the data collection pipeline without introducing shock to the policy gradient.

\subsection{Failure transition and PPO training}
\label{sec:failure_transition}

Suppose the stopping rule fires at step $T_{\mathrm{stop}}$.
ESPO then converts the current prefix into an absorbing failure transition:
\begin{equation}
    r_t = 0 \quad (t < T_{\mathrm{stop}}), \qquad
    r_{T_{\mathrm{stop}}} = r_{\mathrm{fail}}.
\label{eq:failure_reward}
\end{equation}
No further decoding is performed after $T_{\mathrm{stop}}$, and no bootstrap term is
applied beyond the absorbing state.
The resulting trajectory is therefore shorter than the nominal rollout horizon, but
its advantages are still computed by the same GAE recursion in Equation~\eqref{eq:gae}.
In particular, the stopping event induces a negative temporal-difference signal at the
termination point, which is propagated backward to earlier steps through PPO training.

The final training objective remains the standard PPO formulation; ESPO changes the sampled trajectories, not the algebraic form of the policy update. Applying an absorbing penalty exactly at the termination step inherently avoids the pathologies associated with explicit per-step penalties. A state-dependent per-step penalty would introduce a non-stationary reward function that biases the critic and inadvertently incentivizes the policy to collapse its logit spread rather than solve the task. 

\begin{algorithm}[t]
\caption{ESPO Rollout Collection}
\label{alg:ESPO}
\begin{algorithmic}[1]
\REQUIRE Policy $\pi_\theta$, critic $V_\phi$, terminal penalty $r_{\mathrm{fail}}$
\REQUIRE Parameters $\alpha_{\mathrm{s}}, \beta, \varepsilon$, warmup status
\STATE $z \leftarrow 0$; $t \leftarrow 0$; $\mathrm{done} \leftarrow \mathrm{False}$
\WHILE{$t < T_{\max}$ \AND \NOT $\mathrm{done}$}
    \STATE Compute logits $\ell = \log \pi_\theta(\cdot \mid s_t)$; sample $a_t \sim \pi_\theta(\cdot \mid s_t)$
    \STATE $g_t \leftarrow \max_a \ell_a - \ell_{a_t}$ \hfill\COMMENT{Single-step surrogate regret}
    \STATE $\tilde{g}_t \leftarrow \clip\bigl((g_t - \mu_g) / \sqrt{\sigma_g^2 + \delta}, -c, c\bigr)$
        \hfill\COMMENT{Uses frozen batch EMA}
    \STATE $z \leftarrow \alpha_{\mathrm{s}} z + (1 - \alpha_{\mathrm{s}}) \tilde{g}_t$ \hfill\COMMENT{Exponential smoothing}
    \IF{warmup complete \AND $z > \beta \cdot \max\bigl(V_\phi(s_t),\;\varepsilon\bigr)$}
        \STATE $r_t \leftarrow r_{\mathrm{fail}}$; $\mathrm{done} \leftarrow \mathrm{True}$ \hfill\COMMENT{Absorbing failure transition}
    \ELSE
        \STATE $r_t \leftarrow 0$; $t \leftarrow t + 1$ \hfill\COMMENT{Continue decoding}
    \ENDIF
\ENDWHILE
\RETURN trajectory $\tau = (s_0, a_0, r_0, \ldots, s_{T_{\mathrm{stop}}}, a_{T_{\mathrm{stop}}}, r_{\mathrm{fail}})$
\end{algorithmic}
\end{algorithm}

We acknowledge that formulating ESPO as an augmented MDP introduces an inherent objective bias via false positives, terminating locally uncertain but globally recoverable trajectories. However, the aforementioned critic warmup acts as the primary safeguard against this bias early in training, ensuring that valid exploration is not aggressively truncated before the value function is informative. The complete rollout collection procedure is summarized in Algorithm~\ref{alg:ESPO}.

%% file: sections/5_experiments.tex
\section{Experiments}
\label{sec:experiments}

\subsection{Setup}

\paragraph{Models and Benchmarks}

We evaluate DeepSeek-R1-Distill-Qwen-1.5B~\citep{deepseekai2025deepseekr1incentivizingreasoningcapability}
and DeepSeek-R1-Distill-Qwen-7B~\citep{deepseekai2025deepseekr1incentivizingreasoningcapability} models which are trained on DAPO-Math-17k~\citep{yu2025dapo}. The performance is evaluated across three
held-out benchmarks: AIME24~\citep{aime2024dataset}, AMC23~\citep{aimo_amc2023} and MATH500~\citep{lightman2023let}. 

\paragraph{Baselines and Implementation}
We compare ESPO against (1) Base Model; (2) PPO \citep{schulman2017proximal}, the standard token-level actor-critic baseline with full-horizon rollout collection;  and (3) DAPO \citep{yu2025dapo}, which builds on GRPO with dynamic sampling and advantage clipping to improve training stability. The number of rollout is set to 8 and the global batch size is set as 64 for all methods.
All trained under identical data, reward, sampling, and evaluation settings. All algorithms are implemented via verl~\citep{verl} using outcome-based rewards. Moreover, all
experiments were conducted on 8 × H20 GPUs. All detailed training parameters related are provided in Appendix~\ref{app:hparams}. For evaluation,  to maintain results stability, we repeat the
evaluation 32 times and reporting the Pass@1 (averaged over 32 samples). The hyperparameters of inference are
consistently set to a temperature of 1.0, a top-p of 0.7 and a top-k of -1.0.

\input{figures/TABLE_1_main_results_rapo}

\subsection{Main Results}

\textbf{Overall Performance.} Table~\ref{tab:main_results} presents performance across all benchmarks at both model scales. At the 7B scale, ESPO achieves \textbf{85.83\%} on AMC 2023, \textbf{46.28\%} on AIME 2024, and \textbf{87.42\%} on MATH-500, surpassing both PPO and DAPO on every benchmark while consuming only 839.24M cumulative rollout tokens---roughly \textbf{22\% fewer} than PPO and \textbf{19\% fewer} than DAPO. In terms of average accuracy across all three benchmarks, ESPO achieves \textbf{73.17\%}, compared to 71.20\% for PPO ($+$1.97pp) and 71.76\% for DAPO ($+$1.41pp).

\textbf{1.5B Scale Results.} At the 1.5B scale, ESPO likewise demonstrates consistent improvements. It achieves \textbf{71.87\%} on AMC 2023, \textbf{81.53\%} on MATH-500, and an average accuracy of \textbf{59.09\%} across all benchmarks, outperforming both PPO (57.03\%) and DAPO (58.29\%). Although ESPO's AIME 2024 score of 23.87\% is marginally below DAPO (24.37\%), it consumes only \textbf{927.96M} cumulative tokens---significantly less than DAPO (1223.96M, $-$24\%) and PPO (1069.66M, $-$13\%)---while maintaining competitive accuracy, confirming that early-stopping successfully eliminates post-failure token waste without degrading generation quality. Figure~\ref{fig:hero} (right panel) further illustrates training efficiency in terms of cumulative rollout tokens consumed. 

\textbf{Comparison with DAPO.} DAPO, which introduces dynamic sampling and advantage clipping on top of GRPO, generally outperforms standard PPO at both scales, confirming the importance of improved credit assignment for long-horizon reasoning. ESPO further improves upon DAPO on almost all benchmarks, while consistently using fewer rollout tokens. This demonstrates that ESPO is \textbf{complementary} to advantage-estimation improvements: by eliminating uninformative post-failure tokens from the rollout buffer, ESPO provides a cleaner learning signal regardless of the advantage normalization scheme employed.

%% file: figures/TABLE_1_main_results_rapo.tex
\begin{table}[!ht]
\centering
\caption{Main results on math benchmarks on 1.5B and 7B scales. Cumulative Tokens (M) denotes total rollout tokens consumed over 500 training steps. Avg Tokens denotes mean rollout length per trajectory. Avg acc denotes the average accuracy across the three datasets. Best results in \textbf{bold}.}

\label{tab:main_results}
\centering
\resizebox{\textwidth}{!}{%
\begin{tabular}{lcccccc}
\toprule
\multicolumn{7}{c}{\textit{\textbf{DeepSeek-R1-Distill-Qwen-1.5B}}} \\
\midrule
\textbf{Method} & \textbf{AMC23} & \textbf{AIME24} & \textbf{MATH500} & \textbf{Avg Acc} & \textbf{Cumulative Tokens(M)}  & \textbf{Avg Tokens}\\
\midrule
Base Model & $58.28$ & $20.31$ & $74.81$ & 51.13 & - & 5808.00 \\
PPO & $68.43$ & $23.02$ & $79.65$ & 57.03 & 1069.66 & 4178.37\\
DAPO & $70.23$ & $\textbf{24.37}$ & $80.28$ & 58.29 & 1223.96  & 4781.09\\
\textbf{ESPO(Ours)} & $\textbf{71.87}$ & $23.87$ & $\textbf{81.53}$ & \textbf{59.09} & \textbf{927.96} & $\textbf{3624.86}$\\
\midrule
\multicolumn{7}{c}{\textit{\textbf{DeepSeek-R1-Distill-Qwen-7B}}} \\
\midrule
\textbf{Method} & \textbf{AMC23} & \textbf{AIME24} & \textbf{MATH500} & \textbf{Avg Acc} & \textbf{Cumulative Tokens(M)} & \textbf{Avg Tokens}\\
\midrule
Base Model & $78.64$ & $40.13$ &83.36 & $62.04$ & - & 5357.00\\
PPO & $82.94$ & $45.25$ & $85.43$ &71.20 & 1072.40 & 4189.06\\
DAPO & $83.76$ & $45.57$ & $85.95$ &71.76 & 1035.01 & 4043.00 \\
\textbf{ESPO(Ours)} & $\textbf{85.83}$ & $\textbf{46.28}$ & $\textbf{87.42}$ & $\textbf{73.17}$& $\textbf{839.24}$ & $\textbf{3278.30}$\\
\bottomrule
\end{tabular}
}
\end{table}

%% file: sections/6_ablation.tex
\section{Ablation Studies}
\label{sec:ablation}
\input{figures/TABLE_2_ablation}

\Cref{tab:ablation} isolates the contribution of each ESPO component by
removing or replacing it while keeping all other settings fixed.
Experiments are conducted on DeepSeek-R1-Distill-Qwen-7B.

\paragraph{Critic warmup (A vs.\ B).}
Variant B removes the adaptive warmup schedule described in
\Cref{sec:method}.
The gap between A and B (46.3 vs.\ 44.2) is 2.1 points: the
warmup increases performance because the warmup serves as a primary safeguard against objective bias by preventing the algorithm from aggressively truncating valid exploration before the critic becomes informative. This phase ensures that the value-gated threshold $\beta \cdot \max\!\bigl(V_\phi(s_t),\;\varepsilon\bigr)$ is derived from a stabilized value baseline, thereby avoiding spurious early terminations caused by the high variance of a randomly initialized critic.

\paragraph{Terminal failure penalty (A vs.\ C).}
Removing the terminal failure penalty reduces AIME24 by 2.6 points and adds more than 181 average training rollout tokens per trajectory (from 3278 to 3522). This performance drop occurs because treating the truncation as an absorbing failure state with a specific terminal penalty induces a concentrated negative temporal-difference (TD) error. Through training, this more precise negative-signal efficiently propagates backward to earlier steps, providing precise credit assignment.

\paragraph{Value-only vs.\ regret-only stopping (A vs.\ D vs.\ E).}
Variant D stops when $\Vphi(\st) < \tau$ without any regret signal, where $\tau$ is a fixed threshold, relying
solely on the critic.
Variant E stops when the cumulative deviation $z_t > \tau$ without value gating.
Both underperform full ESPO (A), with variant D scoring 44.0 and variant E
scoring 44.8.
Value-only stopping depends entirely on the Critic's absolute scale, which
varies across tasks and training stages.
Regret-only stopping lacks the recovery allowance that the value term provides.
The combination in variant A outperforms either component alone, confirming that
both signals carry complementary information.

\paragraph{Random stop (A vs.\ F).}
Variant F replaces the surrogate regret with a random stop signal:
the rollout stops randomly when generating a trajectory and to control for variables, we set the rate of random truncation to be the same as that in our method.
This achieves 42.4 on AIME24---below all ESPO variants, indicating that randomly reducing rollout tokens without considering the policy's internal confidence or value estimates fails to remove the specific post-failure noise precisely that hampers learning. Furthermore, despite having a similar average rollout length (3342 tokens) to the full ESPO, Variant F exhibits a significant performance gap of 3.9 points, confirming that the benefits of ESPO stem from "where" the trajectories are truncated rather than simply training on shorter sequences.

\section{Analysis}
\begin{figure}[t]
    \centering
    \begin{minipage}[t]{0.49\columnwidth}
        \centering
        \includegraphics[width=\linewidth]{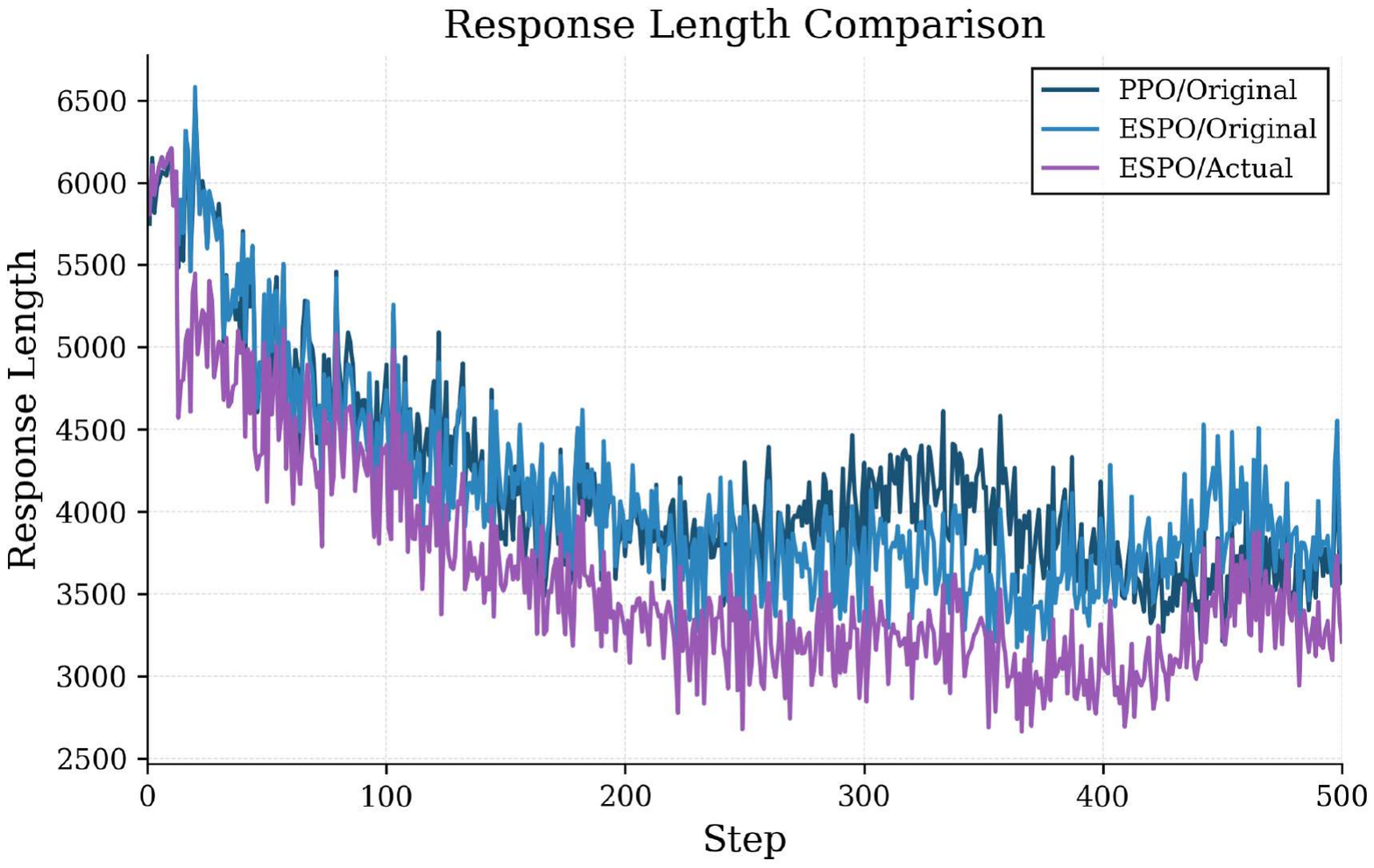}
        \subcaption{Response length over training steps.}
    \end{minipage}
    \hfill
    \begin{minipage}[t]{0.49\columnwidth}
        \centering
        \includegraphics[width=\linewidth]{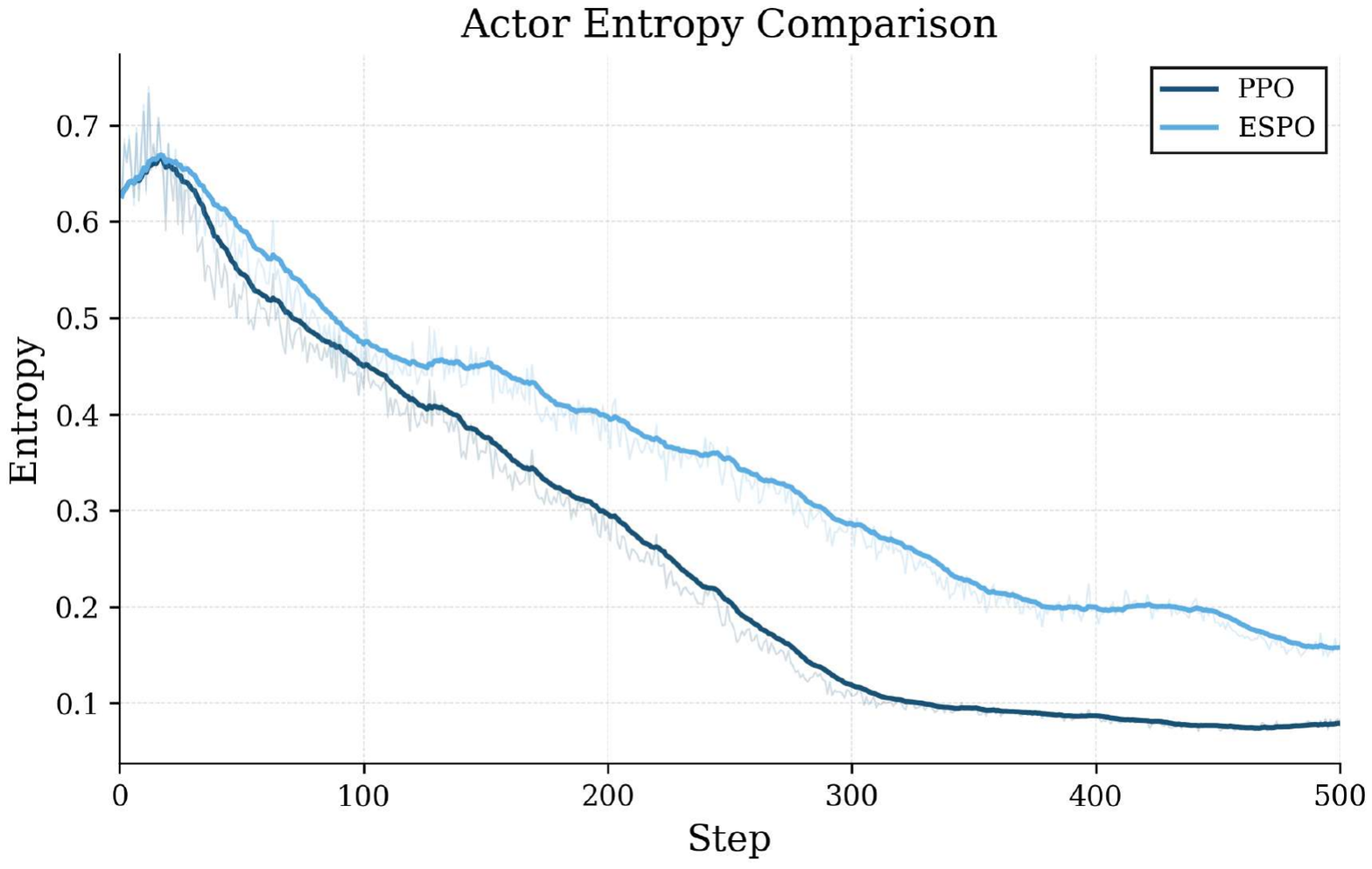}
        \subcaption{Actor entropy over training steps.}
    \end{minipage}

    \vspace{6pt}

    \begin{minipage}[t]{0.49\columnwidth}
        \centering
        \includegraphics[width=\linewidth]{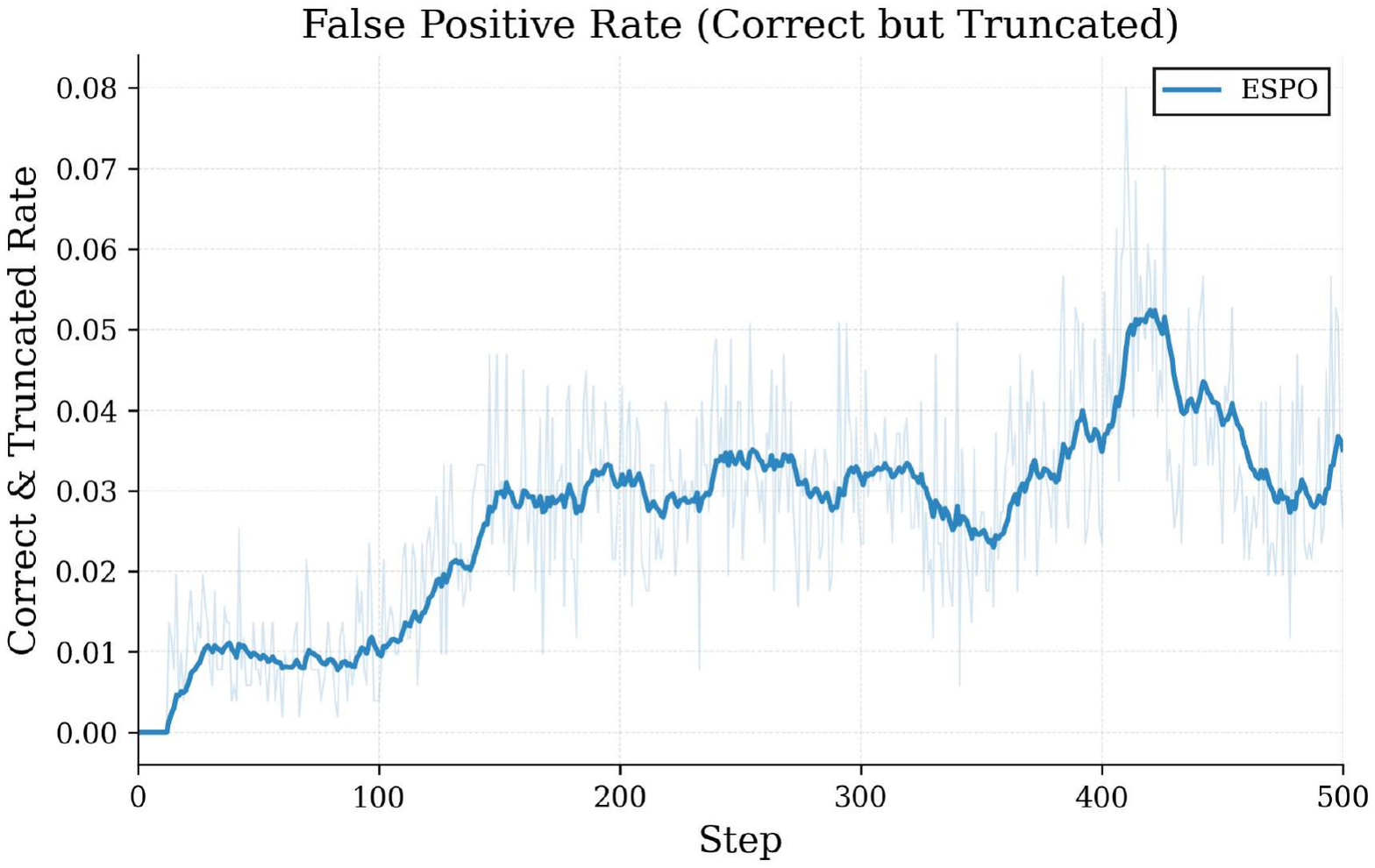}
        \subcaption{False Positive Rate over training steps.}
    \end{minipage}
    \caption{Training dynamics of ESPO vs.\ PPO on DeepSeek-R1-Distill-Qwen-1.5B.
    In~(a), \emph{ESPO/Original} records mean length of all trajectories in ESPO that the model should have generated without executing the early-stopping;
    \emph{PPO/Original} records mean length of all trajectories during training process in PPO;
    \emph{ESPO/Actual} records mean length of all trajectories in ESPO when executing the early-stopping,  including the truncated responses and the non-truncated responses.
    In~(b), ESPO maintains higher entropy throughout training, mitigating premature entropy collapse.
    In~(c), the false positive rate measures the proportion of correct-yet-truncated responses at each step.}
    \label{fig:response_length}
\end{figure}

\subsection{Response Length}

Figure~\ref{fig:response_length}a plots the mean response length curves over training. 
The way we calculate \emph{ESPO/Original} is that during the rollout process, we compute the cumulative regret value as usual. When the first token position satisfying the truncation condition is identified, the sequence is not actually truncated. Instead, we set all mask entries after the truncation point to zero to simulate the truncation. In the subsequent calculation of losses, tokens generated beyond the truncation point are treated as padding tokens and excluded from gradient backpropagation. This allows us to identify the truncated but correct answer sequences. The false-positive rate measuring the proportion of originally correct sequences that are improperly truncated by ESPO is also calculated in this way.

Trajectories that reach natural completion in ESPO (\emph{ESPO/Original}) are closely tracking the mean response length in PPO(\emph{PPO/Original})---while the
actual length of trajectories in ESPO(\emph{ESPO/Actual}) is significantly lower than them, indicating that our method does not destroy the policy's distribution of the response lengths.
At the same time, the accuracy on the validation set also exceeds the baseline method, indicating the positive effectiveness of removing post-failure noise on model training.

\subsection{Policy Entropy and Diversity}

A natural concern is that ESPO's logit-gap signal---which fires when the sampled
token deviates from the policy mode---might suppress exploration by penalizing
low-probability tokens regardless of whether they reflect irrecoverable errors.
To check for this, we track the entropy of $\pitheta(\cdot|s_t)$ over training as stated in Figure~\ref{fig:response_length}b.
ESPO not only does not cause entropy collapse, but also reduces the rate of decrease in the entropy of policy compared to PPO, that is, it opened up the exploration space of the model further.
These controls rule out the hypothesis that ESPO's gains arise from encouraging
greedy/mode-seeking behavior at the expense of exploratory tokens. From another perspective, the stopping strategy of ESPO mainly determines whether to stop by calculating the numerical relationship between value and logit-gap. It does not directly penalize low-probability tokens, so it will not directly lead to a decrease in entropy. Beyond the absence of direct penalty on token probabilities, ESPO actively slows entropy decay by removing a source of spurious gradient signal. In standard PPO, post-failure tokens within a doomed trajectory still receive negative advantages and contribute to the policy gradient, pushing the policy to sharpen its distribution against tokens that were not, in fact, the cause of failure. By truncating these trajectories, ESPO eliminates this misattributed pressure, leaving the policy free to retain probability mass on plausible alternative continuations.

\subsection{False-positive truncation}
We record how
often ESPO truncates trajectories that would have recovered under full rollout in Figure~\ref{fig:response_length}c. We found that, on average in each batch of data, 2.7\% of the trajectories could have yielded correct answers, but are instead truncated, implying an average false-positive truncation rate of 2.7\% in the training process.
These false positives incur a small training cost (the policy misses 2.7\% of
potentially correct trajectories), but this is bounded by the 
improvement over full-horizon PPO, suggesting the benefit of removing
post-failure noise outweighs the cost of occasional early false positives.

%% file: figures/TABLE_2_ablation.tex
\begin{table}[t]
\centering
\caption{Ablation on AIME24 evaluated by avg@32, cumulative tokens and average training rollout tokens per trajectory.
(A) is the full ESPO; all other rows remove or replace one component.
$\uparrow$~higher is better, $\downarrow$~lower is better.}
\label{tab:ablation}
\resizebox{\textwidth}{!}{%
\begin{tabular}{lccc}
\toprule
\textbf{Variant} & \textbf{AIME24\,$\uparrow$} & \textbf{Cumulative Tokens(M)} & \textbf{Avg Tokens\,$\downarrow$} \\
\midrule
  \textbf{(A) Full ESPO (ours)} & \textbf{46.3}  & \textbf{839.24} & \textbf{3278.30} \\
  (B) w/o warmup  & 44.2 &858.37 & 3353.03 \\
  \midrule
  (C) w/o terminal failure penalty  & 43.7 & 901.65 & 3522.09 \\
  (D) Value-only stop ($V_\phi < \tau$, no regret) & 44.0 &1090.05 & 4258.02 \\
  (E) Regret-only stop ($z_t > \tau$, no value gate) & 44.8 &1086.51 & 4244.18 \\
  \midrule
   (F) Random stop & 42.4 &855.59 & 3342.14 \\
\bottomrule
\end{tabular}
}
\end{table}

%% file: sections/7_conclusion.tex
\section{Limitations and Future Work}
\label{sec:limitations}
Our method has limitations when dealing with models that are incorrect but highly confident. In such cases, the surrogate regret approaches zero when the policy is
confidently wrong: a policy that assigns high probability to an
incorrect reasoning branch produces a small logit-gap, delaying detection.
Similarly, as shown in the above error killing rate, ESPO may prematurely trigger the stop condition on a few high-entropy but correct steps, resulting in a false kill. A more refined early stopping strategy can be incorporated into future work to reduce this phenomenon.
The truncation rate needs to be manually adjusted to suit different models and tasks. Future work can explore more adaptive mechanisms to reduce the sensitivity to hyperparameters. Furthermore, extending the stopping criterion to tool-use and multi-turn agentic settings,
where the error may occur across environment steps rather than individual tokens,
is a natural next direction.

\section{Conclusion}
\label{sec:conclusion}

We introduced ESPO, an efficient regret-aware rollout termination method designed to optimize agentic reinforcement learning and LLM reasoning training. By combining a surrogate regret signal derived from the actor's own logit distribution with a dynamic value-gated threshold, ESPO accurately detects and truncates failing trajectories on-the-fly. Applying an implicit failure penalty at the termination step successfully isolates and removes post-failure noise from the policy gradient, completely eliminating the need for auxiliary reward models or human annotations. Extensive evaluations across rigorous mathematical reasoning benchmarks validate the effectiveness of our method. Notably, on DeepSeek-R1-Distill-Qwen-7B, ESPO outperforms the PPO and DAPO baselines on multiple benchmarks, while simultaneously reducing the training rollout tokens cumulatively by more than 20\%. Ultimately, these results demonstrate that ESPO provides a scalable, compute-efficient framework for advancing long-horizon reasoning capabilities in large language models.

%% file: sections/A_appendix.tex
\clearpage
\newpage
\appendix
\section{Training Details}
\label{app:hparams}

For the ESPO training hyperparameters, the learning rate is set to $1 \times 10^{-6}$. The maximum rollout length ($T_{\text{max}}$) is configured to 8192 tokens, alongside a global batch size of 64 and the number of rollout is 8.  Regarding the algorithmic specifics, the failure reward $r_{\text{fail}}$ is set to $-1.0$, the EMA $\alpha_{\text{ema}}$ to 0.99, and the normalisation $\alpha_{\text{s}}$ to 0.9. The initial value of $\beta$ is set to 7.0. A $\beta$ adjustment rate of 0.1 is applied to maintain a target termination rate of 0.25. The baseline methods maintain the same settings, such as the global batch size. Additionally, for DAPO, the clip\_ratio\_low is set to 0.2 and the clip\_ratio\_high is set to 0.28 as stated in \citep{yu2025dapo}.

\begin{table}[h]
\centering
\caption{ESPO hyperparameters.}
\label{tab:hparams}
\begin{tabular}{lc}
\toprule
\textbf{Hyperparameter} & \textbf{Value} \\
\midrule
Failure reward $\rfail$ & $-1.0$ \\
EMA $\alphasm_{EMA}$ & $0.99$ \\
Normalization $\alphasm_{s}$ & $0.9$ \\
Initial $\betath$ & $7.0$ \\
Minimum $\betath_{min}$ & $0.0$ \\
$\betath$ adjustment rate & $0.1$ \\
Target termination rate & $0.25$ \\
Max rollout length $\Tmax$ & $8192$ \\
Learning rate & $1 \times 10^{-6}$ \\
The Value's $\varepsilon$ & $0.2$ \\
\bottomrule
\end{tabular}
\end{table}

\section{Adaptive Critic Warmup Details}
\label{app:warmup}

During warmup, the stopping criterion is disabled and the critic is updated
using only the base PPO objective. Warmup exits early if the absolute value of critic's loss is less than 0.5 or the difference between adjacent step's values is less than 0.1 for three
consecutive steps, indicating the critic has converged.
If critic's loss does not meet the convergence criterion after 10\% of total training
steps, warmup ends unconditionally and the stopping criterion activates
to avoid indefinitely deferring the ESPO mechanism.

\begin{figure}[!ht]
    \centering
    \begin{minipage}[t]{0.47\columnwidth}
        \centering
        \includegraphics[width=\linewidth]{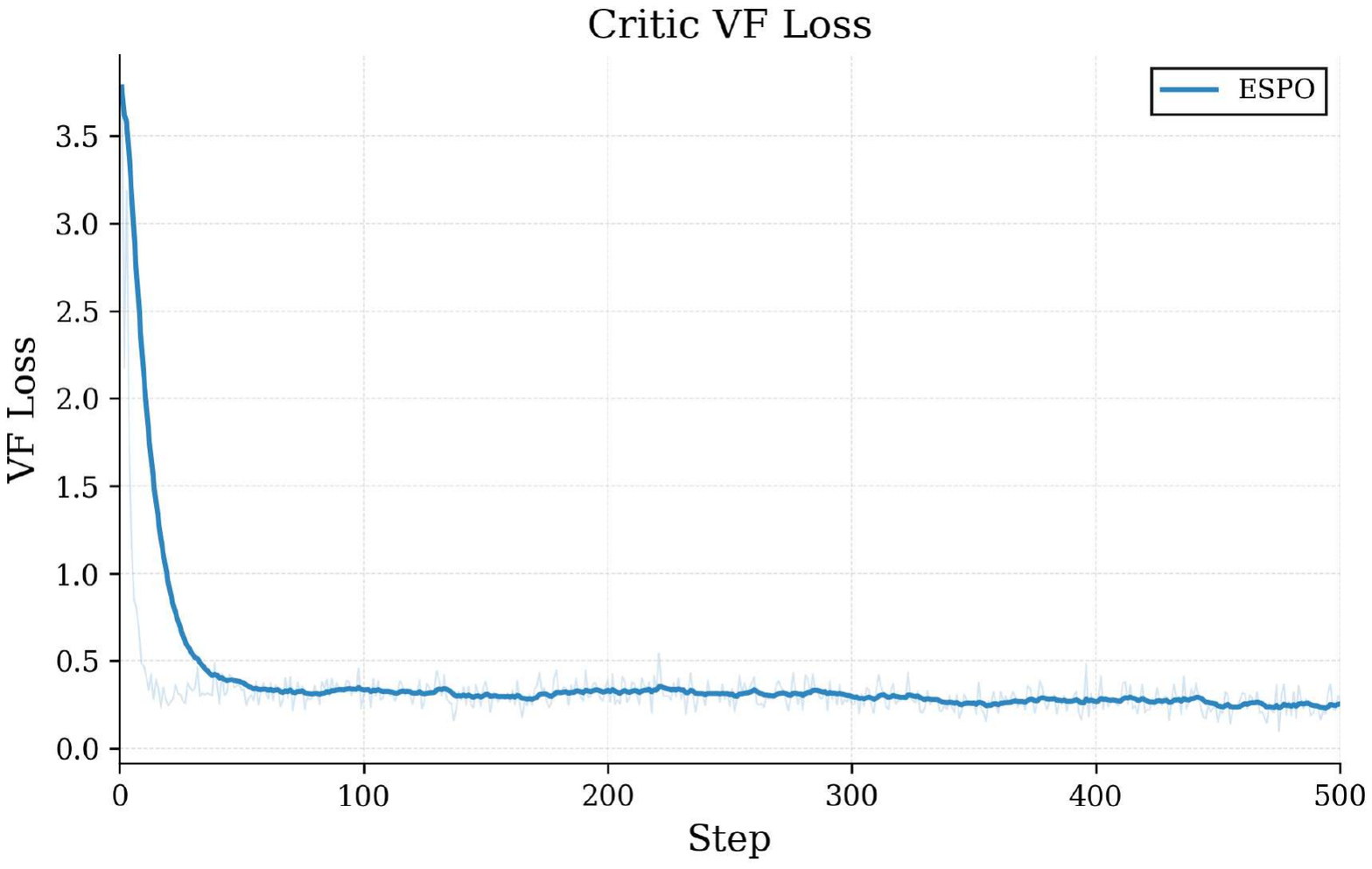}
        \subcaption{Critic loss on DeepSeek-R1-Distill-Qwen-1.5B}
    \end{minipage}
    \hfill
    \begin{minipage}[t]{0.47\columnwidth}
        \centering
        \includegraphics[width=\linewidth]{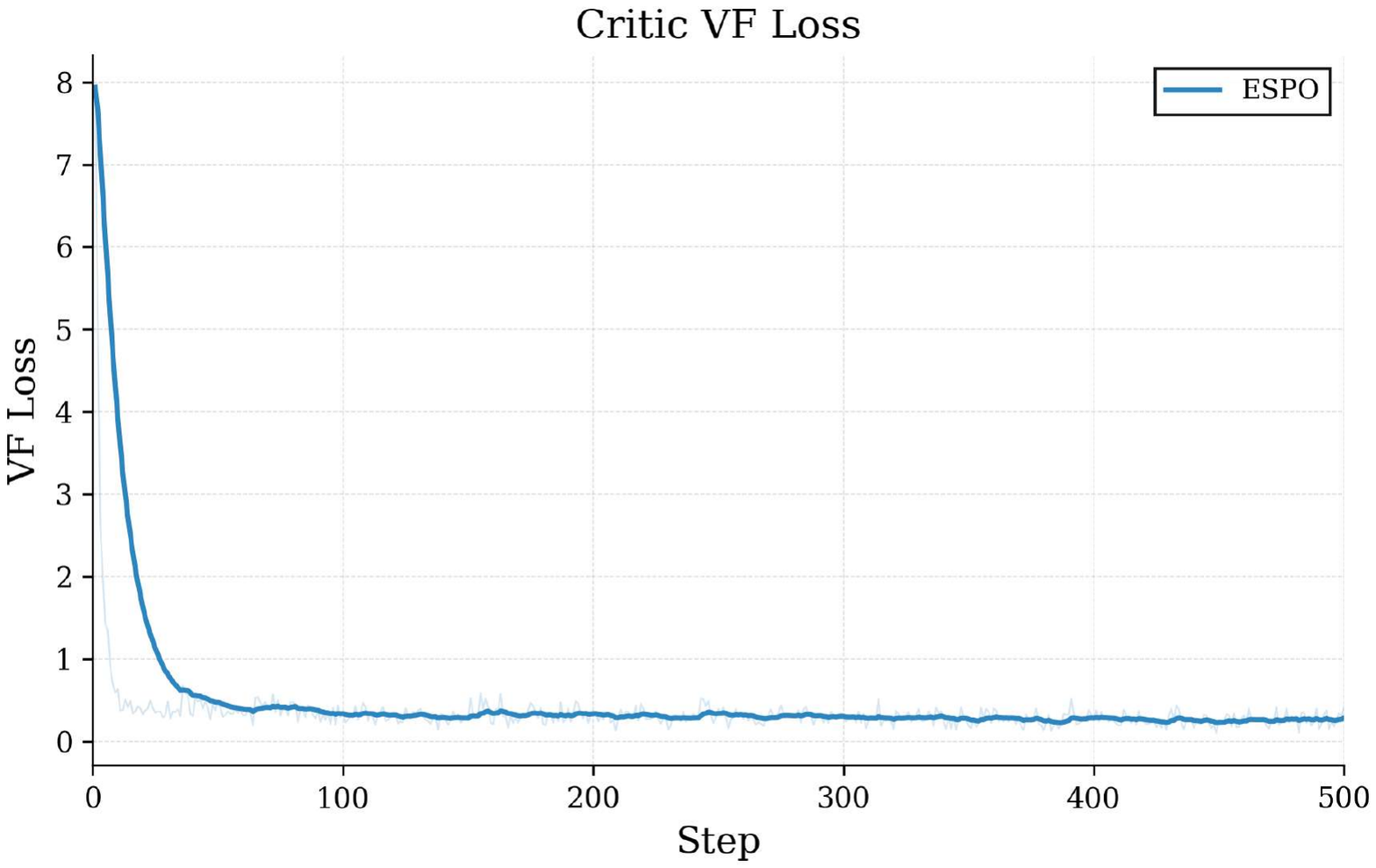}
        \subcaption{Critic loss on DeepSeek-R1-Distill-Qwen-7B}
    \end{minipage}
    \caption{%
        The left figure illustrates the critic loss for DeepSeek-R1-Distill-Qwen-1.5B, while the right figure depicts the critic loss for DeepSeek-R1-Distill-Qwen-7B.}
    \label{fig:criticloss}
\end{figure}